\documentclass[11pt,a4paper]{article}
\usepackage{arabtex}
\usepackage[T1]{fontenc}
\usepackage[utf8]{inputenc}
\usepackage{authblk}

\usepackage[hyperref]{emnlp2020}
\usepackage{times}
\usepackage{multirow}
\usepackage{amsmath}
\usepackage{amssymb}
\usepackage{subcaption}
\usepackage{tipa}
\usepackage{tikz}
\usepackage{pgfplots}
\usepackage{latexsym}

\usepackage{etoolbox}
\usepackage{todonotes}
\usepackage{microtype}
\usepackage{comment}
\usepackage{wasysym}

\aclfinalcopy 
\def\aclpaperid{***} 

\newcommand{\ipa}{\textipa} 

\newcommand*{\affaddr}[1]{#1} 
\newcommand*{\affmark}[1][*]{\textsuperscript{#1}}
\newcommand*{\email}[1]{\texttt{#1}}
\makeatletter
\newcommand{\printfnsymbol}[1]{%
	\textsuperscript{\@fnsymbol{#1}}%
}
\makeatother

\title{The Persian Dependency Treebank Made Universal}
\author{%
	Mohammad Sadegh Rasooli\affmark[1]\thanks{~~Rasooli and Safari equally contributed in the conversion and experimentation process. Rasooli and Moloodi equally contributed in the linguistic design of conversion rules and manual investigation of conversions.}, Pegah Safari\affmark[2]\printfnsymbol{1}, Amirsaeid Moloodi\affmark[3]\printfnsymbol{1}, Alireza Nourian\affmark[4]\\
	\affaddr{\affmark[1] Department of Computer and Information Science,  University of Pennsylvania, Philadelphia, PA, USA}\\
	\affaddr{\affmark[2] Faculty of Computer Science and Engineering,
		Shahid Beheshti University, Tehran, Iran }\\
		\affaddr{\affmark[3] Department of Foreign Languages and Linguistics, Shiraz University, Shiraz, Iran }\\
		\affaddr{\affmark[4] Sobhe, Tehran, Iran }\\
	\affaddr{\affmark[1]\email{rasooli@seas.upenn.edu}}, \affaddr{\affmark[2]\email{p\_safari@sbu.ac.ir}},\\ \affaddr{\affmark[3]\email{amirsaeid.moloodi@shirazu.ac.ir}}, \affaddr{\affmark[4]\email{nourian@sobhe.ir}}\\
}

\date{}

\begin{document}
\maketitle
\begin{abstract}
We describe an automatic method for converting the Persian Dependency Treebank \cite{rasooli-etal-2013-development} to Universal Dependencies. This treebank contains $29107$ sentences. Our experiments along with manual linguistic analysis show that our data is more compatible with Universal Dependencies than the Uppsala Persian Universal Dependency Treebank~\cite{seraji-etal-2016-universal}, and is larger in size and more diverse in vocabulary. Our treebank brings in a labeled attachment F-score of $85.2$ in supervised parsing. Our delexicalized  Persian-to-English parser transfer experiments show that a parsing model trained on our data is $\approx$2\% absolutely more accurate than that of  \newcite{seraji-etal-2016-universal} in terms of labeled attachment score.\footnote{The treebank is publicly available in \url{https://github.com/UniversalDependencies/UD_Persian-PerDT/tree/dev}.}
\end{abstract}

\section{Introduction}
In recent years, there has been a great deal of interest in developing universal dependency treebanks~\cite{mcdonald-etal-2013-universal,rosa-etal-2014-hamledt,nivre2020universal}.  The main goal of the Universal Dependencies project~\cite{nivre2020universal} is to develop a consistent linguistic annotation scheme in different levels from tokenization to syntactic dependency relations. As a result, the majority of annotation discrepancies disappear, and the resulting dataset  facilitates several cross-lingual natural language processing tasks including  part-of-speech transfer~\cite{tackstrom-etal-2013-token},  syntactic transfer~\cite{naseem-etal-2010-using,mcdonald2011multi,ammar-etal-2016-many,zhang-etal-2019-cross}, and probing~\cite{tenney-etal-2019-bert,hewitt-manning-2019-structural}. Starting with 10 treebanks in 2015, there are 163 treebanks in version 2.6 (May 2020) including the Uppsala Persian Treebank \cite{seraji-etal-2016-universal}.

Persian (aka Farsi) is a pro-drop morphologically rich language with a high degree of free word order and a unique light verb construction~\cite{karimi2011separability}. Despite its importance, it still suffers from lack of sufficient annotated data. The Uppsala Universal treebank~\cite{seraji-etal-2016-universal}  is currently the only publicly available universal treebank for Persian. It is a valuable resource based on news genre, and has been used as a testbed in previous work~\cite{zeman-etal-2018-conll,chi2020finding}. Among other non-universal treebanks, the Persian dependency treebank (PerDT)~\cite{rasooli-etal-2013-development} is significantly larger than \cite{seraji-etal-2016-universal} (29K  vs. 6K sentences), and its sentences are sampled from contemporary Persian texts in different genres (as opposed to only news genre). 

\begin{figure}
	\centering
	\includegraphics[]{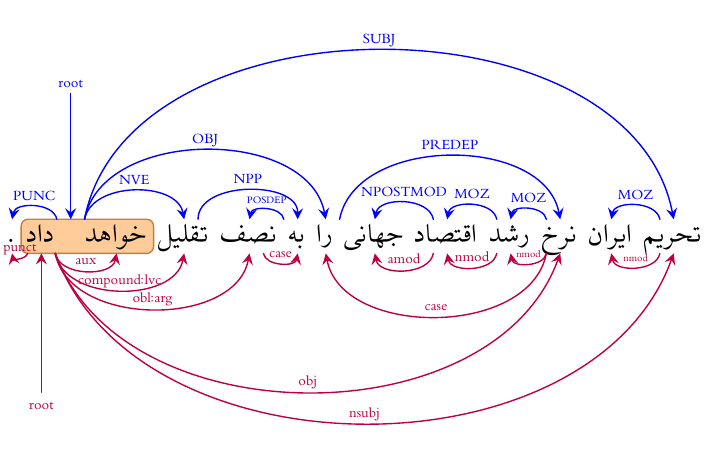}
	\caption{An example of our automatic conversion.  The universal labels are shown at the bottom of words.}\label{fig:convert}
\end{figure}

In this paper, we propose an automatic method for converting PerDT~\cite{rasooli-etal-2013-development} to Universal Dependencies (An example of such conversion is shown in Figure~\ref{fig:convert}). After a thorough analysis of dependency relations in the treebank, we design different mapping rules to generate trees with universal relations. This process involves a series of steps including unifying tokenization, part-of-speech tags, named-entity recognition, and finally mapping dependencies. The mapping for many relations are not necessarily one-to-one, and we have to deal with peculiar cases that are specific to certain structures in modern Persian. Therefore, our approach is neither a blind one-to-one mapping, nor an expensive and time-consuming manual process. We empirically show that our annotations are more compatible with the Universal guidelines via learning a delexicalized transfer model with more than $2\%$ absolute difference in labeled attachment score.  The summary of our contributions is as following:
\begin{itemize}
	\setlength\itemsep{0em}
	\item We propose an automatic annotation conversion process with manual care of special cases.  We develop a new Persian Universal Treebank with 29107 sentences. This is in contrast to the treebank of \newcite{seraji-etal-2016-universal} that contains 5997 sentences.  
 \item We develop a modified and corrected version of  PerDT with the Universal tokenization scheme. Moreover, the new release resolves various tagging errors in the original dataset. Most of these corrections are made by manually fixing annotation errors flagged by our mapping pipeline.
\end{itemize}

\section{Related Work}\label{sec:related}
There has been a great deal of interest in designing and developing Persian dependency treebanks~\cite{pouramini2007annotation,seraji2012bootstrapping,seraji-etal-2014-persian,seraji-etal-2016-universal,rasooli2011syntactic,rasooli-etal-2013-development,ghayoomi2014converting}. Among them, the Uppsala UD treebank~\cite{seraji-etal-2016-universal} is the only treebank with Universal Dependencies. We have found some caveats in the Uppsala Universal Treebank~\cite{seraji-etal-2016-universal}. This causes annotation discrepancies in some frequently used dependency relations such as \emph{compound:lvc}, \emph{cop}, \emph{csubj}, \emph{fixed}, \emph{obl}, and \emph{xcomp} (see \S\ref{sec:upp} for more details).

We primarily focus on converting the Persian dependency treebank (PerDT)~\cite{rasooli-etal-2013-development}. PerDT has been used in previous studies for Persian dependency parsing~\cite{khallash-etal-2013-empirical,feely-etal-2014-cmu,nourian-etal-2015-importance,pakzad2016improved}. It has been extended to other representations including semantic roles~\cite{mirzaei-moloodi-2016-persian} and discourse~\cite{mirzaei-safari-2018-persian}. It is also included in the HamleDT collection~\cite{rosa-etal-2014-hamledt}.

\section{Approach}
In the conversion process, we have noticed several key differences between PerDT and UD. We decompose the conversion process into 3 steps: 1) tokenization, 2) part-of-speech mapping, 3) systematic changes to PerDT, and 4) dependency relation mapping. In this section, we briefly describe the mentioned steps.

\subsection{Tokenization}\label{sec:tok}
There are two key differences in PerDT tokenization from UD: 1) Multiword inflections of simple verbs in Persian are grouped as one word with spaces in between parts following the deterministic rules from \cite{rasooli2011unsupervised}.  We follow the guidelines in \cite[Table 3]{rasooli-etal-2013-development} to find out the main verb and make other parts an ``aux'' dependent of the main verb. We introduce the ``AUX'' part-of-speech tag and ``aux'' dependency relation (``aux:pass'' for passive verbs) in this UD tokenization scheme. 2) Clitics are only detached from words in cases for which they play an object or verbal role. Other clitics are pronominal clitics attached to nouns, prepositions, pronouns and adjectives. By looking at the word lemma, we recover those pronouns, and detach them, and assign their heads to the closest nominal word with the ``MOZ'' (\emph{Ezafe}) dependency label.

\subsection{POS Mapping}\label{sec:pos}
This  is the most straightforward step except for proper nouns. We could only discover a small portion of them by finding noun phrases with an identifier (IDEN POS tag for words such as ``Dr.'' or ``Mr."). In addition to mapping the IDEN POS to PROPN, we use a recent BERT-based Persian named-entity tagger~\cite{taher2020beheshti} to recover additional proper nouns.  The tagger can find 7 different entities including date, location, money, organization, percent, person and time. We only consider the \emph{person} and \emph{location} entities, and manually revise the results to add missing entities, foreign words and the name of months. Table~\ref{tab:pos_mapping} shows the mappings. 

\begin{center}

\begin{table}[t!]

\small \centering
\begin{tabular}{ l  l    l  }
	\hline \hline
			\setlength{\tabcolsep}{3pt}
{\bf PerDT}    & {\bf Condition}  &{\bf  UD  } \\  \hline
V      &  & VERB  \\ \hline
\multirow{2}{*}{N}      &  NER=False & NOUN  \\
    &  NER=True & PROPN  \\ \hline
SUBR   &  & SCONJ \\ \hline
CONJ   &  & CCONJ \\ \hline
ADV    &    & ADV   \\ \hline
\multirow{2}{*}{ADJ}    &NER=False & ADJ   \\
   &  NER=True  & PROPN  \\ \hline
PR     &  & PRON  \\  \hline
PUNC   &  & PUNCT \\  \hline
ADR    &  & INTJ  \\ \hline
IDEN   &  & PROPN \\  \hline
\multirow{3}{*}{ PART}   & Word=\setfarsi\novocalize \< rA > & ADP \\
  &Word$\in${\tiny \{\setfarsi\novocalize \<xwb>, \setfarsi\novocalize \<Axr>\}} & INTJ\\  
  & Otherwise & PART \\ \hline
PREM   &  & DET   \\ \hline
\multirow{2}{*}{PRENUM}   & Cardinal & NUM \\
 & Ordinal  & ADJ  \\ \hline
PREP  &  &  \multirow{2}{*}{ADP} \\
 POSTP &  &    \\ \hline
 \multirow{2}{*}{POSTNUM}   & Cardinal & NUM \\
 & Ordinal  & ADJ  \\ \hline
PSUS   &  & INTJ \\  	\hline \hline
\end{tabular}
\caption{Mapping rules for part-of-speech tags. }
\label{tab:pos_mapping}
\end{table}

\end{center}

\begin{center}

\begin{table}[t!]

\scriptsize \centering
\setlength{\tabcolsep}{3pt}
\begin{tabular}{| l | l  |  l | l | }
	\hline
{\bf PerDT}    & {\bf Precondition}  & {\bf Pre-action} &{\bf  UD  } \\ \hline 
ACL &  & \multirow{2}{*}{CMR (mark)~[fig.\ref{fig:cmr}]} &  \multirow{2}{*}{ccomp}   \\
PRD & & &  \\ \hline
\multirow{2}{*}{ADV}      & ?$\rightarrow$\{ADJ,ADV\} &  &   advmod \\
& Otherwise &  &  obl  \\ \hline
AJCONJ   &   & \multirow{5}{*}{Conj rotation~[fig.\ref{fig:conj_rotate}] } & \multirow{5}{*}{conj} \\  
AVCONJ   &  &  &  \\
NCONJ   &  &  &  \\    
PCONJ   &  &  &  \\  
VCONJ   &   & &  \\  \hline
AJUCL &  & CMR (mark)~[fig.\ref{fig:cmr}] &advcl   \\ \hline
 & ?$\rightarrow$\{ADP\} & CMR (case)~[fig.\ref{fig:cmr}] &   obl \\
  & ?$\rightarrow$\{NUM\} &  &   nummod  \\ 
APOSTMOD& ?$\rightarrow$\emph{nominal} &  &   nmod  \\
APREMOD &  ?$\rightarrow$\{ADJ\} &  &   amod  \\
 & ?$\rightarrow$DET  &  &   det \\
&  ?$\rightarrow$\{ADV\} &  &   advmod  \\ \hline
ADVC &  & \multirow{5}{*}{CMR (case)~[fig.\ref{fig:cmr}] }  &   \multirow{5}{*}{obl:arg} \\
NEZ &  &  & \\
AJPP & & & \\
VPP & &  &   \\ 
VPRT & &  &   \\ \hline
APP  &   & & appos \\  \hline
\multirow{2}{*}{COMPPP} &  $\exists$ dep & CMR (case)~[fig.\ref{fig:cmr}]  &  case  \\
 & Otherwise & & fixed \\ \hline 
NEZ &  &  & \\ \hline
ENC   &   & & \multirow{4}{*}{compound:lvc} \\  
NE   &  &  &  \\  
NPRT   &  &  CMR (case)~[fig.\ref{fig:cmr}] &  \\ 
NVE   &   & &  \\  \hline
LVP & &  &  compound:lv \\ \hline 
\multirow{2}{*}{NCL} & ?$\rightarrow$SCONJ & CMR (mark)~[fig.\ref{fig:cmr}] &  \multirow{2}{*}{acl}   \\
& Otherwise &   &  \\ \hline
MESU & & Dep $\rightarrow$Head (Flip) & nmod \\ \hline
\multirow{2}{*}{MOS} & AUX$\rightarrow$? &  Dep $\rightarrow$Head (Flip)  & cop\\ 
  &  Otherwise & & xcomp  \\ \hline
    \multirow{3}{*}{MOZ}      &  ?$\rightarrow$NOUN &  \multirow{3}{*}{CMR (case)~[fig.\ref{fig:cmr}] } &  nmod \\
  & ?$\rightarrow$ADJ &   &  amod \\ 
  &  Otherwise &    & advmod  \\ \hline
  \multirow{3}{*}{NADV}      &  ?$\rightarrow$NOUN &  \multirow{3}{*}{CMR (case)~[fig.\ref{fig:cmr}] } &  nmod \\
  & ?$\rightarrow$ADJ &   &  amod \\ 
  &  Otherwise &    & advmod  \\ \hline
NPOSTMOD &  & &  amod \\  \hline
 \multirow{2}{*}{NPP}      &\{NVE$|$ENC\}$\rightarrow$? & NPP rotation~[fig.\ref{fig:npp_rotation}]  & obl:arg  \\
& Otherwise &  CMR (case)~[fig.\ref{fig:cmr}] &  nmod  \\ \hline
 \multirow{3}{*}{NPREMOD}      & ?$\rightarrow$DET  &  &   det\\
 &  ?$\rightarrow$ cardinal &   &  nummod \\ 
 & Otherwise &    & amod \\ \hline
\multirow{2}{*}{OBJ} & $\exists$ OBJ2 sib. &  &   iobj \\
& Otherwise  &  &   obj \\  \hline
OBJ2 &   &  &   obj \\  \hline
\multirow{2}{*}{PARCL} & ?$\rightarrow$CCONJ  &\multirow{2}{*}{Dep $\rightarrow$Head (Flip)} &   conj  \\
& Otherwise  &   &   parataxis \\  \hline
PART   &   & & mark \\  \hline
PUNC   &   & & punct \\  \hline
\multirow{2}{*}{PROG}      &  Active$\rightarrow$?  & & aux  \\
&  Passive$\rightarrow$?  & & aux:pass  \\ \hline
ROOT   &   & & root \\  \hline
\multirow{2}{*}{SBJ}      & Active$\rightarrow$? &  & nsubj  \\
 &  Passive$\rightarrow$? &  & nsubj:pass  \\ \hline
 TAM   &  & &  xcomp \\  \hline
\multirow{3}{*}{VCL }&  Modal verb$\rightarrow$? & Dep $\rightarrow$Head (Flip) & aux \\ 
& $\exists$ MOS,$\nexists$ SUBJ sib. &  \multirow{2}{*}{CMR (mark)~[fig.\ref{fig:cmr}]}  & csubj \\
  & Otherwise & &  ccomp \\    \hline
   \multirow{5}{*}{PREDEP}   & NUM$\rightarrow$?& & advmod \\ 
 &  NOUN$\rightarrow$PRON  & & dislocated \\
   &  ?$\rightarrow$CCONJ & & cc  \\
   &  ?$\rightarrow$NOUN & & obl  \\
 &   \emph{Last mapping}  & & advmod \\   \hline
  \multirow{4}{*}{ POSDEP}   &  NOUN$\rightarrow${\tiny \{\setfarsi\novocalize \<hm, nyz>\}} & & dep  \\
 &  ?$\rightarrow$NOUN & & obl  \\
    &  ?$\rightarrow$CCONJ & & cc  \\
    & \emph{Last mapping} & & advmod\\  \hline
  
\end{tabular}
\caption{Mapping rules for dependencies. PerDT labels are described in \newcite[Table 2]{rasooli-etal-2013-development}. First \emph{Preconditions} (2nd column) should satisfy. Afterwards,  \emph{Preactions} (3rd column) are applied  before applying the UD conversions (4th column). These preactions are depicted in Figure~\ref{fig:all_rules}.}
\label{tab:dep_mapping}
\end{table}

\end{center}

\subsection{Systematic Changes to PerDT}\label{sec:sys_change}
Before starting to convert the treebank~\cite{dadegan2012persian}, we have made the following systematic changes to PerDT:
\begin{figure}[t!]
\centering
\includegraphics[width=0.4\textwidth]{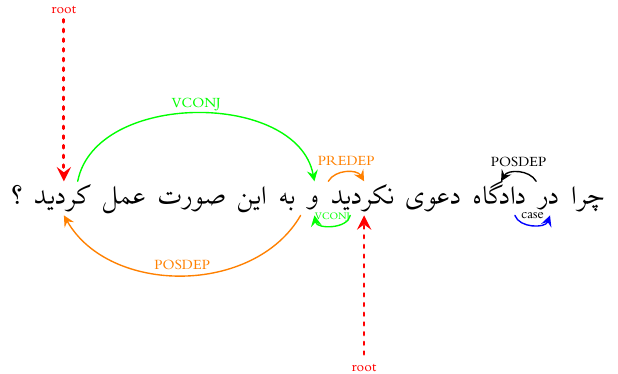}
\caption{The result of applying rotations of conjunctions for the sentence \emph{Why didn't you defend in the court {\bf and} acted like that?}. In this example, case rotation for preposition is also shown.}\label{fig:rules_result}
\end{figure}

\begin{itemize}
	\item We convert the order of verbal conjunctions in the original data. In PerDT, verbal conjunctions are conventionally attached from the end to the beginning~\cite{dadegan2012persian}.\footnote{Examples in \url{https://bit.ly/2Mfz1iH}} We find this convention unintuitive and reverted the order of conjunctions.  Figure~\ref{fig:rules_result} shows an example of such rotation.

	\item Words such as ``billion'', ``million'', ``thousand'' are tagged as nouns.\footnote{Examples in \url{https://bit.ly/2Y105Yv}} This might be due to the fact that these words can be inflected as plurals while number should not be inflected in Persian. We believe that a better tagging decision for these words is number since their inflection as plurals is due to a special kind of  \emph{zero derivation} or \emph{conversion}  numbers to nouns in particular contexts~\cite{booij2012grammar}.
	\item PerDT assumes that all inflections of {\small ``\setfarsi\novocalize \<^sdn>''} \ipa{[Sod\ae n]}  is passive and its lemma is  {\small ``\setfarsi\novocalize \<krdn>''} \ipa{[k\ae d\ae n]}. We have changed this assumption and use the superficial lemma for those instances. The decision makes our data similar to the annotations of \newcite{seraji-etal-2016-universal}.
	
\end{itemize}

\begin{table}[t!]
	\small
	\centering
	\begin{tabular}{|l |l|c |l |}
		\hline
		\multicolumn{2}{|c|}{Correction Type}           & \#    & \multicolumn{1}{c|}{\%}    \\ \hline
		& Systematic & 3694  & \multicolumn{1}{c|}{0.762} \\ \cline{2-4} 
		\multirow{-2}{*}{Lemma}            & Others      & 59    & 0.012                                              \\ \hline
		& Systematic & 529   & 0.109                                              \\ \cline{2-4} 
		\multirow{-2}{*}{POS}              & Others      & 298   & 0.061                                              \\ \hline
		& Systematic & 3693  & 0.762                                              \\ \cline{2-4} 
		\multirow{-2}{*}{FPOS} & Others      & 90    & 0.018                                              \\ \hline
		& Systematic & 27407 & 5.658                                              \\ \cline{2-4} 
		\multirow{-2}{*}{Dependency head}             & Others      & 967   & 0.199                                              \\ \hline
		& Systematic & 18516 & 3.823                                              \\ \cline{2-4} 
		\multirow{-2}{*}{Dependency label}       & Others      & 656   & 0.135                                              \\ \hline
		\multicolumn{2}{|l|}{Word Form}                 & 39    & \multicolumn{1}{c|}{0.008} \\ \hline
	\end{tabular}
	
	\caption{Statistics of PerDT corrections. By systematic, we mean deterministic corrections such as verbal conjunctions (see \S\ref{sec:sys_change} for details). }\label{tab:uni_correct}
	
\end{table}

Table~\ref{tab:uni_correct} shows the statistics of changes that we have made to the data including systematic changes and fixes to incorrect annotations.

\subsection{Dependency Relation Mapping} 
PerDT contains 43 syntactic relations for which many of them cannot easily map to UD. Moreover, conjunctions in PerDT are arranged from the beginning of the sentence to the end in chain-style manner. More importantly, compared to UD scheme for which content words are considered as heads, PerDT assigns prepositions as the head of prepositional phrases and auxiliary verbs as the head of sentences.

Before applying the conversion rules, we label words that are not well-edited and typed as more than one token as \emph{goeswith}. We then label proper noun phrases that are not syntactically compositional as \emph{flat:name}. We also analyze complex numbers as \emph{flat:num} and their coordinating conjunctions as \emph{cc} dependent of each following word. Afterwards we follow the rules in Table~\ref{tab:dep_mapping}. As depicted in the Table, there are conditions that should be satisfied before applying a conversion, and some actions such as flipping a head with its dependent are needed  before certain mappings. Finally, we label the few remaining undecided dependencies as \emph{dep}.

\begin{figure*}[t!]
	\centering
	\scalebox{1.0}{

	\vspace{-0.5cm} 
		\hspace{-0.5cm} 
	\begin{subfigure}[]{0.5\textwidth}
		\centering
\includegraphics[]{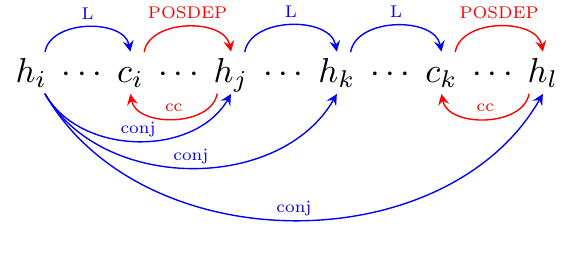}
\caption{Conj Rotation} \label{fig:conj_rotate}
\end{subfigure}
\begin{subfigure}[]{0.25\textwidth}
	\centering
\includegraphics[]{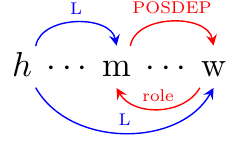}
\caption{CMR (\textcolor{red}{role})} \label{fig:cmr}
\end{subfigure}
\begin{subfigure}[]{0.25\textwidth}
	\centering
\includegraphics[]{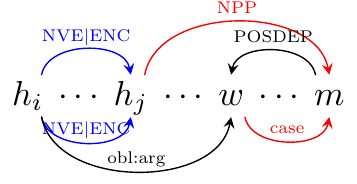}
\caption{NPP rotation} \label{fig:npp_rotation}
\end{subfigure}
}
\caption{A graphical depiction of rotation rules used in this work (see Table~\ref{tab:dep_mapping} for their use cases).}\label{fig:all_rules}
\end{figure*}

\section{Experiments and Analysis}

The general statistics of our data vs. the Uppsala treebank~\cite{seraji-etal-2016-universal} are shown in Table~\ref{tab:sizes}. We observe that our data is superior in many aspects including size and diversity compared to the Uppsala Treebank~\cite{seraji-etal-2016-universal}. The most important fact about PerDT is that its sentences are intentionally sampled in order to cover almost all verbs from the Verb Valency Lexicon~\cite{rasooli2011syntactic} leading to  3.9 times more verb lemmas than the Uppsala Treebank. Table~\ref{tab:univ_freq} shows the counts of each dependency label in the converted Data.

\begin{table}[t!]
	\small
	\centering
\setlength{\tabcolsep}{5pt}

	\begin{tabular}{ l  l  c c c c c }
		  
			\multirow{2}{*}{Part} & 	\multirow{2}{*}{Data} & 	\multirow{2}{*}{Sen.\#} & 	\multirow{2}{*}{Tok.\#} & \multicolumn{3}{c}{Types\#} \\
			 & & & & Word  & Lemma & Verb \\ \hline \hline
	\multirow{2}{*}{Train}  & UDT& 4798 & 122K & 13.9K & 6.7K   & 1226\\ 
	& {\bf  Ours} & {\bf 26196} & {\bf 459K} & {\bf 34.9K}&  {\bf 20.7K}& {\bf 5275}\\  \hline \hline
		\multirow{2}{*}{Dev}  & UDT & 599 & 15K & 3.9K &  2.0K & 278 \\ 
	&  {\bf  Ours}  & {\bf 1456}& {\bf 26K}& {\bf 7.0K}&{\bf 5.2K}  &{\bf 1427} \\  \hline \hline
		\multirow{2}{*}{Test}  & UDT & 600 & 16K & 3.9K & 3.1K & 385 \\  
	&  {\bf  Ours} & {\bf 1455}& {\bf 24K}& {\bf 6.7K}&{\bf 5.1K}  & {\bf 1671}\\ \hline  \hline
		\multirow{2}{*}{All}  & UDT & 5997 & 154K & 15.8K & 7.6K & 1387 \\  
	&  {\bf  Ours}  & {\bf 29107} &  {\bf 509K} & {\bf 36.7K} &  {\bf 21.6K}& {\bf 5413}\\ 
	\hline 
	\end{tabular}
\caption{Statistics of our data vs. UDT~\protect\cite{seraji-etal-2016-universal} in different data splits. }
\label{tab:sizes}
\end{table}

\begin{table}[t!]

\small \centering

	\begin{tabular}{l c c }
		\hline \hline
		Label & Frequency & \%  \\  	\hline \hline
  case & 71118 & 14.1 \\
  conj & 23739 & 4.7 \\
  acl & 10034 & 1.9 \\  
  obl & 30737 & 6.1 \\
   punct & 44336 & 8.8 \\
    cop & 6366 & 1.2 \\  
  det & 10273 & 2 \\
   advmod & 9158 & 1.8 \\
    aux:pass & 822 & 0.1 \\  
  nmod & 59442 & 11.6 \\
   appos & 1059 & 0.2 \\
    aux & 12886 & 0.16 \\  
  amod & 22576 & 4.4 \\
   compound:lvc & 32339 & 6.4 \\
    nsubj:pass & 822 & 0.1 \\  
  nsubj & 27181 & 5.4 \\
   name:flat & 7899 & 1.5 \\
    dep & 2035 & 0.4 \\  
  cc & 21300 & 4.2 \\
   root & 29107 & 5.8 \\
    advcl & 4228 & 0.8 \\  
  obj & 19999 & 3.9 \\
   xcomp & 4920 & 0.9 \\
    parataxis & 82 & 0.01 \\  
  ccomp & 6945 & 1.3 \\
   obl:arg & 21510 & 4.2 \\
    flat:num & 607 & 0.1 \\  
  nummod & 5459 & 1 \\
   mark & 11982 & 2.3 \\
    fixed & 144 & 0.02 \\  
  compound:lv & 439 & 0.08 \\
   csubj & 682 & 0.1 \\
    vocative & 174 & 0.03 \\  
  compound & 42 & 0.008 \\
   iobj & 6 & 0.001 \\
    dislocated & 1 & 0.0001 \\  	\hline \hline
 
	\end{tabular}
	
	\caption{Frequency of each universal label  the converted dataset.}\label{tab:univ_freq}
\end{table}

\paragraph{Supervised Parsing}
We evaluate the resulting data by training UDPipe V.2~\cite{udpipe:2017} along with the pre-trained fastText~\cite{grave-etal-2018-learning} embeddings on our data. We also evaluate our models on the Uppsala treebank~\cite{seraji-etal-2016-universal}.  Table~\ref{tab:main_results} shows the parsing results using a trained model on our data and the Uppsala Treebank evaluated by the CoNLL 2018 shared task evaluation scripts~\cite{zeman-etal-2018-conll}. It is worth noting that the goal of this evaluation is not to show which dataset brings in better parsing accuracy: it is clear that the bigger the dataset is, the higher the accuracy can be. Our goal is to show that there is a significant performance difference between the models trained on the two datasets by using the exact same training pipeline. As shown in  Table~\ref{tab:main_results}, we see that there is a huge tagging and parsing performance difference when we move across the datasets. There are two possible reasons: domain mismatch, and annotation discrepancy. Our analysis show that annotation discrepancy plays an important role here. As described in \S\ref{sec:upp}, there are some core incompatibilities between the Uppsala treebank~\cite{seraji-etal-2016-universal} and Universal Dependencies guidelines.  Our detailed analysis shows that most of cross-dataset errors come from errors in \emph{nmod}, \emph{obl}, \emph{fixed}, and \emph{xcomp}. This is in fact consistent with our manual analysis in~\S\ref{sec:upp}.

\begin{table}[t!]
	\small
	\centering
	\begin{tabular}{l  c c c|| c c c }
		 
		Test Data & \multicolumn{3}{c}{PerDT (Ours)} &  \multicolumn{3}{c}{\newcite{seraji-etal-2016-universal}} \\ \hline 	\hline 
		ID tagger &  $\times$ & \multicolumn{2}{c||}{\checkmark} &    $\times$ & \multicolumn{2}{c}{\checkmark}   \\ 
		ID parser  & $\times$ & $\times$ & \checkmark& $\times$ & $\times$ & \checkmark \\ \hline
Tokens & 99.9 & \multicolumn{2}{c||}{{\bf  99.99}}  & 100 &  \multicolumn{2}{c}{100.0}  \\ 
	Words & 99.1 & \multicolumn{2}{c||}{{\bf  99.64}}  & {\bf 99.7} & \multicolumn{2}{c}{99.59}   \\ 
	UPOS        & 82.9 & \multicolumn{2}{c||}{{\bf  96.11}}  & 81.9 & \multicolumn{2}{c}{{\bf  95.75}}   \\ 
	Lemmas     &  80.7 & \multicolumn{2}{c||}{{\bf  96.20}}  & 90.2 & \multicolumn{2}{c}{{\bf  89.55}} \\ 
	UAS       & 71.2 & 71.2& {\bf 88.4} & 69.5 & 69.8 & {\bf 83.5 }\\ 
	LAS       & 64.4 & 62.6 & {\bf 85.2} &  62.1 & 61.0 & {\bf  79.4} \\ 
	CLAS        & 59.9 & 59.3 & {\bf 81.6} & 56.9 & 56.1 & {\bf 74.8 }\\ 
	MLAS       &49.5 & 54.5 & {\bf  78.9} & 46.0 & 53.9 & {\bf 73.0 } \\ 
	BLEX        & 44.6& 56.9 & {\bf 78.2} & 52.1 &49.0 & {\bf 65.5 } \\  
	\end{tabular}
\caption{Parsing results based on the CoNLL shared task 2018~\protect\cite{zeman-etal-2018-conll} evaluation. ID stands for in-domain for which the same training set is used for training a UDPipe model~\protect\cite{udpipe:2017}}
\label{tab:main_results}
\end{table}

\paragraph{Delexicalized Model Transfer}
One  way to verify our claim about increased consistency of our UD conversion with the UD guidelines  is to learn a transfer model. In this setting, we follow the delexicalized parser transfer approach which have been extensively used in previous work~\cite{zeman-resnik-2008-cross,mcdonald2011multi,tackstrom-etal-2012-cross}. We  sample the same number of tokens as of \newcite{seraji-etal-2016-universal} from PerDT. Afterwards, we delexicalize both of the treebanks, and learn a parser using the Yara Parser~\cite{rasooli2015yara}. We train two models  with 15 epochs and evaluate them on the delexicaled test set of the Universal English Web Treebank~\cite{silveira-etal-2014-gold}. The model trained on PerDT significantly  outperforms the other model by 2\% both in unlabeled and labeled attachment score (47.31 vs 45.37 UAS, 38.59 vs. 36.45 LAS). This is a strong indicator that our data is more compatible with the UD annotations.

\section{Conclusion}
We have introduced our approach in making PerDT~\cite{rasooli-etal-2013-development} universal.  During this process, we have faced different challenges such as annotation errors in the original data, tokenization inconsistencies, lack of named entities, part-of-speech and dependency label mapping. Due to automatic conversions and potential annotation errors in the original treebank, there is always a chance of some annotation incompatibilities between our treebank and the Universal guidelines. Therefore, we cannot claim that our conversion is perfect. However, our experiments have shown that our data is more compatible with the Universal Dependencies guidelines than the Uppsala treebank~\cite{seraji-etal-2016-universal}.  

\section*{Acknowledgments}
We would like to thank Morteza Rezaei-Sharifabadi for his efforts in getting copyright permissions and making our treebank publicly available. We also thank Daniel Zeman for helping us in the process of releasing this treebank.

\appendix 
\section*{Appendix}


\section{Problems in the Universal Annotations of the Uppsala Universal Treebank}\label{sec:upp}
We briefly mention some of the problems in the Uppsala Universal Treebank~\cite{seraji-etal-2016-universal}:
\begin{itemize}
	  \setlength\itemsep{0em}
	\item \newcite{seraji-etal-2016-universal} does not determine the \emph{csubj} label in their analysis. For example, in ``\textcolor{purple}{\ipa{lAzem}} \textcolor{green}{\ipa{P\ae st}} \ipa{Pu}  {\bf \textcolor{red}{\ipa{beres\ae d}}}'' (it \textcolor{green}{is} \textcolor{purple}{necessary} for him {\bf \textcolor{red}{to arrive}}), it is obvious that what comes after ``\textcolor{green}{\ipa{P\ae st}}'' is the clausal subject of the adjectival sentence predicate ``\textcolor{purple}{\ipa{lAzem}}''. A simple syntactic test supports this viewpoint: one can convert the clausal complement ``\ipa{Pu}  {\bf \textcolor{red}{\ipa{beres\ae d}}}'' to a noun phrase ``\textcolor{red}{\ipa{resid\ae n}}\ipa{-e} \ipa{u}'' (his \textcolor{red}{arrival}). The new phrase plays the \emph{nsubj} role of the sentence. Therefore, the clausal complement of the sentence should be \emph{csubj}. Our converted data contains $682$ cases of \emph{csubj}.
\item  \newcite{seraji-etal-2016-universal} considers prepositional and possessive complements of adjectival heads as \emph{nmod} and \emph{nmod:poss} respectively. Their analysis clearly stands in contradiction to UD annotation guideline in which \emph{nmod} is used just for dependents of a nominal head.  \emph{obl} is much better suited for these cases.
\item  \newcite{seraji-etal-2016-universal} consider {\small```\setfarsi\novocalize \<^sdn>''} \ipa{[Sod\ae n]} (to become) as copula. What UD asserts under the \emph{cop} (copula) label is that \emph{``the equivalents of to become are not copulas despite the fact that traditional grammar may label them as such.''} Instead, it should be deemed as a verbal predicate and its second complement as \emph{xcomp}. 
\item {\small`` \setfarsi\novocalize \<pydA krdn>''}  (``\ipa{peydA k\ae rd\ae n}'')  and {\small`` \setfarsi\novocalize \<.hA.sl krdn>''}  (``\ipa{hAsel k\ae rd\ae n}'') are considered as two-word light verbs~\cite{2017moloodi}. We consider the non-verbal part as the first part of the two-word light verb, and use the \emph{compound:lv} label for it ($439$ cases in PerDT). However, \newcite{seraji-etal-2016-universal} annotate the nonverbal elements of these complex predicates as \emph{obj} and considers  ``\ipa{peydA}'' as a nonverbal element.
\item \emph{iobj} label is absent in \cite{seraji-etal-2016-universal}, most likely due to the low frequency of this syntactic relation. Our converted treebank contains $6$ cases of  \emph{iobj}.
\item Proper nouns are not labeled in \cite{seraji-etal-2016-universal}. Ours covers proper nouns (more than $23K$ tokens).

\end{itemize}

\bibliography{refs}
\bibliographystyle{acl_natbib}
\end{document}